\documentclass[11pt]{article}
\usepackage{emnlp2019}
\usepackage{multirow}
\usepackage{times}
\usepackage{latexsym}
\usepackage{graphicx}
\usepackage{footnote}
\usepackage{enumitem}
\makesavenoteenv{tabular}
\makesavenoteenv{table}

\usepackage{url}

\aclfinalcopy 

\title{NewB: 200,000+ Sentences for Political Bias Detection}

\author{Jerry Wei \\
  Dartmouth College / Hanover, NH \\
  {\tt jerry.weng.wei@gmail.com} \\}

\date{}

\begin{document}
\maketitle
\begin{abstract}
We present the Newspaper Bias Dataset (NewB), a text corpus of more than 200,000 sentences from eleven news sources regarding Donald Trump. 
While previous datasets have labeled sentences as either liberal or conservative, NewB covers the political views of eleven popular media sources, capturing more nuanced political viewpoints than a traditional binary classification system does. 
We train two state-of-the-art deep learning models to predict the news source of a given sentence from eleven newspapers and find that a recurrent neural network achieved top-1, top-3, and top-5 accuracies of 33.3$\%$, 61.4$\%$, and 77.6$\%$, respectively, significantly outperforming a baseline logistic regression model's accuracies of 18.3$\%$, 42.6$\%$, and 60.8$\%$. 
Using the news source label of sentences, we analyze the top n-grams with our model to gain meaningful insight into the portrayal of Trump by media sources.
We hope that our dataset (\url{https://github.com/JerryWeiAI/NewB}) will encourage further research in using natural language processing to analyze complex political biases.
\end{abstract}

\section{Introduction}

Newspaper articles are often biased \cite{Pinholster1067, MediaBias} and reflect the political leaning of their news source \cite{Media=Divide}.
In recent American politics, the actions of current U.S. President Donald J. Trump have polarized the American people \cite{DonaldDivides} and are thus an exemplary topic of political bias in media. Conservative news sources are more likely to report on Trump's actions favorably, whereas liberal media outlets tend to portray Trump's actions in a more negative light \cite{ElectionBias}. 

\begin{table}[ht]
\small
 \setlength{\abovecaptionskip}{5pt}
\centering
\begin{tabular}{ l | c c c } 
 \hline
     Source (Political Bias) & $N$    & $w_{avg}$ & $\vert$V$\vert$ \\ 
 \hline
 \hline
 Newsday (L) & 24,000 & 23 & 66,366\\ 
 New York Times (L) & 24,000 & 18 & 88,982\\ 
 Cable News Network (L) & 24,000 & 25 & 53,096\\ 
 Los Angeles Times (L) & 24,000 & 25 & 70,073\\ 
 Washington Post (L) & 24,000 & 25 & 67,648\\ 
 Politico (N) & 24,000 & 28 & 203,725\\ 
 Wall Street Journal (C) & 24,000 & 18 & 60,677\\ 
 New York Post (C) & 24,000 & 23 & 50,182\\ 
 Daily Press (C)  & 24,000 & 24 & 60,607\\ 
 Daily Herald (C)  & 24,000 & 25 & 53,515\\ 
 Chicago Tribune (C)  & 24,000 & 24 & 67,953\\ 
 \hline
 Combined & 264,000 & 23 & 362,649\\
 \hline
\end{tabular}
\caption{Summary statistics of our presented Newspaper Bias Dataset (NewB). $N$: number of sentences. $w_{avg}$: average number of words per sentence. $\vert$V$\vert$: vocabulary size. L: liberal source. N: neutral source. C: conservative source. Political leanings of newspapers were retrieved from \textit{Media Bias/Fact Check}.}
\label{table: 1}
\end{table}

The publication of newspaper articles online has generated a large amount of text data conducive for empirical text analysis techniques. 
We collect tens of thousands of newspaper articles regarding Donald Trump and compile them into a sentence-level \textbf{New}spaper \textbf{B}ias dataset called NewB.  
Whereas previous political bias datasets label sentences into the broad categories of liberal and conservative, NewB is labeled by newspaper source and captures more intricate political biases and viewpoints than a binary labeling system can. 
Our work has the following contributions:

\begin{itemize}[leftmargin=*]
  \setlength\itemsep{-0.3em}
    \item We present NewB, a dataset of 200,000$+$ sentences from articles regarding Donald Trump. 
    Previous work has generically categorized sentences as either liberal or conservative, but NewB is labeled by and covers the political ideologies of eleven newspapers, allowing for more nuanced analysis of political bias.
    \item We train deep learning classifiers to predict the source of a given sentence.
    While predicting the source from just a single sentence is challenging, we find that the patterns discovered by training on such a large text corpus shed significant insight on the biases of newspapers. 
\end{itemize}

\section{Related Work}
Many previous works have used natural language processing for political language and implicit bias analysis. 
For news content, Gentzkow and Shapiro \shortcite{MediaSlant} measured the political ``slant" of news articles by identifying similarities between the articles' language and that of Congressional representatives, and Iyyer et al. \shortcite{PolitIdeologyRNN} used recursive neural networks for political ideology detection on two sentence-level datasets. 
In terms of social media, Rao et al. \shortcite{TwitUserOrientation} classified Twitter users as liberal or conservative through their tweets, and Tumasjan et al. \shortcite{TwitElectionPred} predicted election results using Twitter messages.
Moreover, Rao and Spasojevic \shortcite{MessageDemOrRep} used word embeddings and recurrent neural networks to classify social media messages from Twitter, Facebook, and Google+ as leaning Democratic or Republican. 

Related work also includes several  datasets on political bias and subjectivity that have been released publicly. 
Thomas et al. \shortcite{Convote} released a dataset called Convote for detection of support for or opposition against proposed legislation from Congressional floor-debate transcripts, and Gross et al. \shortcite{IdeologicalBookCorpus} generated a comprehensive dataset of sentences from books and magazine articles labeled as liberal, conservative, or neutral based on the political leaning of the author.

While these works are a solid foundation in political text analysis, our dataset takes a novel approach. Previous political bias and subjectivity datasets have straightforward binary labels but are time-intensive to annotate and often subjective to annotators' comprehension of the text. Our dataset, on the other hand, is labeled by indisputable ground truth classes- the origins of article- that are used for training and inference. Furthermore, while binary labels for political bias are usually sufficient for simple classification tasks, texts often have more intricate political bias than can be captured by binary labels. By having labels that span eleven news outlets, our dataset differentiates between and captures a wider range of political ideologies, allowing for more comprehensive bias analysis. Our paper is the first to our knowledge to propose a newspaper origin classification task and our dataset at least one order of magnitude larger than related datasets (Table 2).

\begin{table}[ht]
\small
 \setlength{\abovecaptionskip}{5pt}
\centering
\begin{tabular}{ l | c c c } 
 \hline
     Dataset & Task & $N$    & $c$\\ 
 \hline
 \hline
 IBC \footnote{\cite{IdeologicalBookCorpus}} & Political Bias & 4,062   & 3 \\
 Convote \footnote{\cite{Convote}} & Legislation Support & 3,857   & 2\\
 SUBJ \footnote{\cite{SubjectivityDataset}}  & Sentence Subjectivity & 10,000  & 2\\
 \textbf{NewB}    & \textbf{News Origin} & \textbf{264,000} & \textbf{11}\\ 
 \hline
\end{tabular}
\caption{Comparison of the NewB dataset with other benchmark political bias and subjectivity datasets. $N$: dataset size. $c$: number of classes.}
\label{table: 2}
\end{table}

\begin{figure*}[ht]
\centering
 \setlength{\abovecaptionskip}{5pt}
    \includegraphics[width=0.8\linewidth]{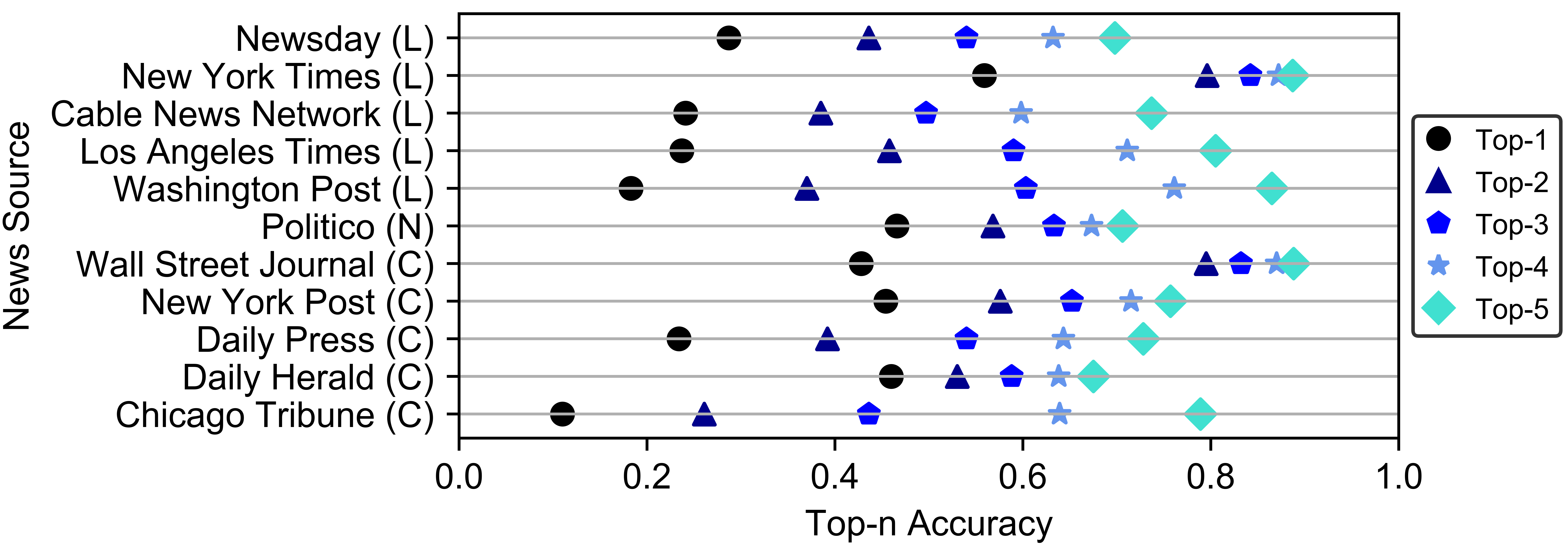}
    \caption{Top-n accuracies by news source for a recurrent neural network trained on NewB.}
    \label{fig:dotplot}
\end{figure*}

\section{The NewB Dataset}

For data collection, we focus on articles that contain the search term ``Donald Trump'' as a topic of contention between liberal sources and conservative sources.
We download texts from the ProQuest US News Stream Database that were written by journalists from the following media outlets: Newsday, New York Times, Cable News Network, Los Angeles Times, Washington Post, Politico, Wall Street Journal, New York Post, Daily Press, Daily Herald, and Chicago Tribune.
We select these news sources based on a cross-reference between available sources on ProQuest US News Stream and sources whose political leaning is shown on \textit{Media Bias/Fact Check} (the largest website that classifies news sources on the political spectrum), and further balance them to include exactly five liberal sources, five conservative sources, and one neutral source. 

In terms of data pre-processing, we split all articles at the sentence level and balance the class distribution to 24,000 sentences per class, for a total of 264,000 distinct sentences across 11 classes. 
Then, we assign all sentences their respective origins as ground truth labels. 
For each class, we randomly partition 1,000 sentences into an independent test set; the total size of our test set is 11,000 sentences. 
We use the remainder of the sentences for model training. 
Our dataset is separated such that there is a standard testing set, which can be useful for future comparisons between classifiers trained on NewB.
News sources and their political leanings, as well as summary statistics of NewB can be found in Table 1. 
Of note, the origin of a text does not necessarily directly illicit bias for a given article (e.g. a New York Times article may report positively about President Trump even though New York Times articles generally report negatively about him). 

\section{Experimental Setup}

\subsection{Text Classification Models}
To establish baselines for performance on our news source classification dataset, we implement the following text classifiers:

\begin{itemize}[leftmargin=*]
  \setlength\itemsep{-0.3em}
    \item \textbf{Logistic Regression (LR):} Logistic Regression is the simplest neural network architecture shown to satisfy the universal approximation theorem \cite{LogisticRegression}.
    \item \textbf{Convolutional Neural Network (CNN):} Convolutional neural networks, which are commonly used in computer vision, have been shown to achieve high performance on text classification tasks \cite{CNN>RNN}. We implement a CNN with a single 1D convolutional layer, followed by max pooling and a dense layer.
    \item \textbf{Recurrent Neural Network (RNN):} Recurrent neural networks are commonly used in language processing because they are suitable for processing sequential data. We use an RNN with two bidirectional layers of LSTM cells \cite{LSTMRNN}. 
\end{itemize}

\subsection{Model Training and Evaluation}
Our models take variable-length sentences as inputs and return softmax output vectors of length eleven representing the predicted confidence for each news sources, which are compared to ground truth labels represented as one-hot encoded vectors.
When inputting sentences into the model, we tokenize each sentence at the word-level and convert each word into a vector using 300-dimensional distributed embeddings trained on the Common Crawl database with the GloVe method \cite{GloVe}. 
Each model was trained using a five percent cross-validation split until convergence was determined using early stopping.

\begin{table}[ht]
 \setlength{\abovecaptionskip}{5pt}
\centering
\small
    \begin{tabular}{ l || c c c c c} 
    \hline
    \multirow{2}{4em}{Model} & \multicolumn{5}{c}{Top-n Accuracy ($\%$)}\\
     \cline{2-6}
    & $n$$=$$1$ & $n$$=$$2$ & $n$$=$$3$ & $n$$=$$4$ & $n$$=$$5$\\ 
     \hline
    LR  & 18.3 & 32.1 & 42.6 & 52.2 & 60.8\\
    CNN & \textbf{34.0} & 50.3 & \textbf{61.5} & 70.0 & 77.4\\
    RNN & 33.3 & \textbf{50.6} & 61.4 & \textbf{70.5} & \textbf{77.6}\\
    \hline
    \end{tabular}
\caption{Top-n accuracies for each model. LR: logistic regression. CNN: convolutional neural network. RNN: recurrent neural network.}
\label{table:3}
\end{table}
\section{Results}
We measure the performances of various baseline models on our dataset and analyze top n-grams with one of our models to gain insight on how Trump is portrayed by various media sources.

\subsection{Evaluation Metrics}
For each model, we calculate top-1,2,3,4,5 accuracies per class, which are shown in Table \ref{table:3}. 
Both the CNN and RNN models significantly outperform the logistic regression baseline, likely as a result of their abilities to account for sequential data. 
With respect to the difficulty of predicting the origin of a news source given just a single sentence, the CNN and RNN achieved respectable accuracies of 34.0\% and 33.3\% respectively, well above a random guessing accuracy of 9.09\%. 

\begin{figure}[ht]
 \setlength{\abovecaptionskip}{5pt}
    \includegraphics[width=7.5cm]{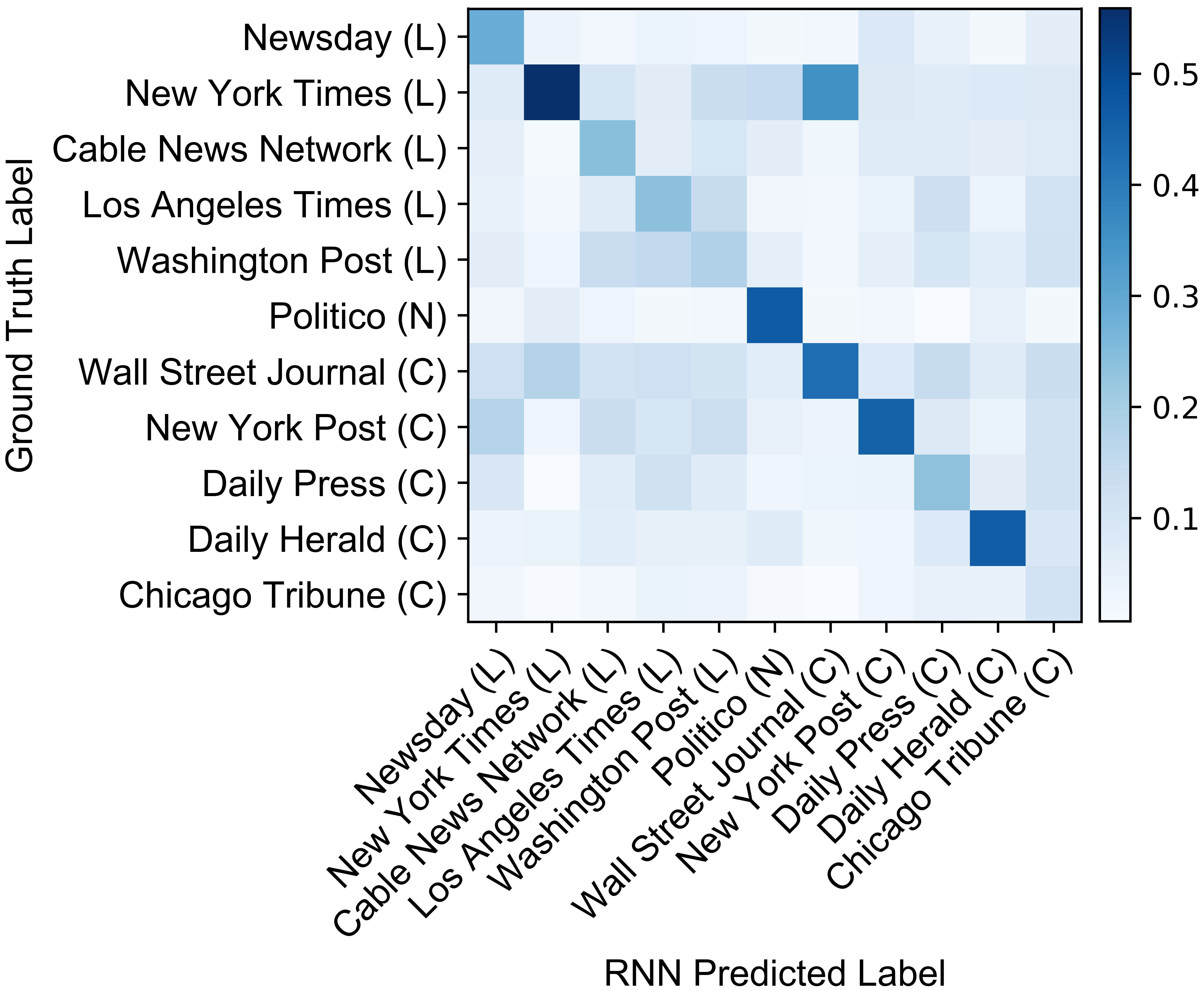}
    \caption{Confusion matrix of predicted and ground truth news sources for a recurrent neural network (RNN) trained on NewB.}
    \label{fig:heatmap}
\end{figure}

In Figure \ref{fig:dotplot}, we show the top-n accuracy per class for the RNN. 
New York Times and Politico were the easiest to classify.
This is may be because Politico uses a larger variety of words than the other classes as shown in its larger vocabulary size, and New York Times has the shortest average sentence length (Table 2).
Furthermore, we display the confusion matrix of the RNN's predicted labels in the form of a heatmap in Figure \ref{fig:heatmap}.
Of note, there was a high false positive rate for predicting Wall Street Journal on New York Times sentences, likely because both newspapers tended to have short sentences, with an average length of only 18 words per sentence.

\subsection{Five-gram Analysis}
\begin{figure}[ht]
 \setlength{\abovecaptionskip}{5pt}
    \includegraphics[width=7.5cm]{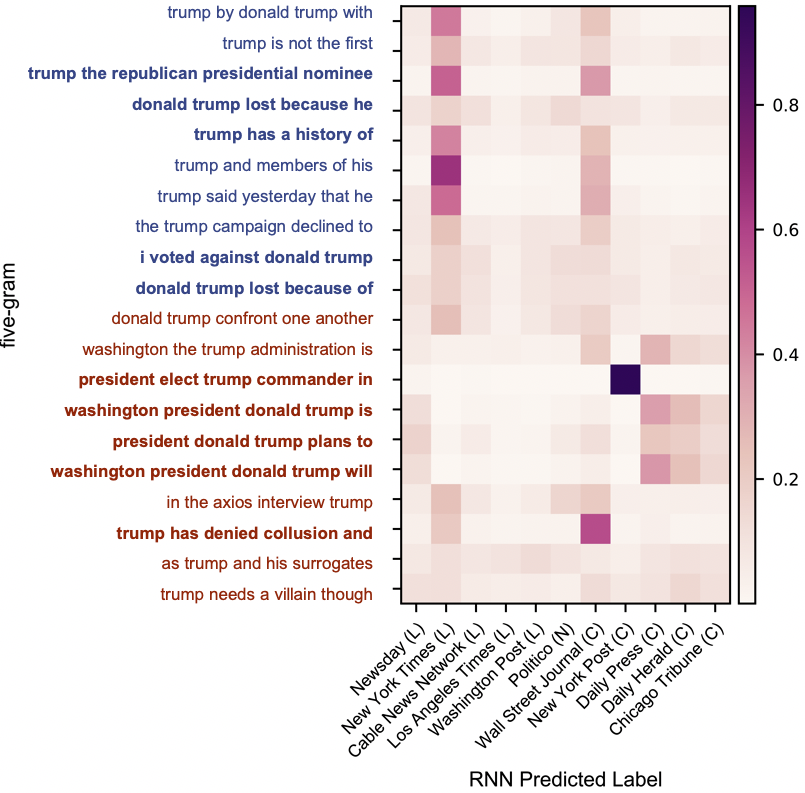}
    \caption{Recurrent neural network's predicted labels for most significant liberal and conservative five-grams. Liberal five-grams (top half) are in blue, and conservative five-grams (bottom half) are in red. Significant n-grams are bolded. }
    \label{fig:n-grams}
\end{figure}

For analysis of our dataset, we find the most significant five-grams for liberal and conservative newspapers by retrieving the most used five-grams that appear only in sources of that bias. 
We input these five-grams into the RNN and display the outputs in a heatmap, as shown in Figure \ref{fig:n-grams}.

Liberal n-grams tend to use loaded words that convey a negative image of Trump.
An in-depth investigation of top five-grams revealed that \textit{trump has a history of} was used negatively ten out of twelve times, and \textit{trump the republican presidential nominee} was used negatively fifteen out of eighteen times, suggesting that liberal sources reported negatively on and distanced themselves from Trump.
Furthermore, the fact that these phrases are from liberal backgrounds is obvious: \textit{the trump campaign declined to comment} is consistent with the fact that the Trump campaign is less likely to respond to liberal news sources, and \textit{i voted against donald trump} is more likely to be found in an article from a liberal source.

On the other hand, conservative newspapers tend to portray Trump in a more positive light. 
They often refer to Donald Trump with titles of respect such as \textit{President} and \textit{Commander-in-Chief}.
Moreover, they frequently supported his claims regarding his alleged collusion with Russia, emphasizing that \textit{trump has denied collusion}.
Overall, our model generally predicted liberal news sources for liberal n-grams, with an emphasis on New York Times, while conservative n-grams were predicted to originate from conservative news sources, with more varied predictions.

\section{Conclusion}
In this paper, we have presented NewB, a newspaper bias dataset of sentences regarding Donald Trump.
While previous works have typically classified political bias broadly (e.g. liberal and conservative), there are often more nuanced biases involved in political texts.
Our main contribution is a dataset that captures these complex biases by labeling each sentence with its news source origin. 
We have shown that substantial insights on the political viewpoints of media sources can be identified by models trained on this data, without the use of manually annotated labels required by previous datasets. 
In terms of applications, a model trained on NewB could potentially be used as an internet browser extension to better inform readers of the biases present in online newspaper articles.

While NewB sets up the foundation for more complex political bias analysis, our current work has not used comprehensive pre-training strategies such as BERT \cite{BERT} and only identifies some of the political tendencies encompassed by this large text corpus.
Important future work remains in developing new methods to translate the learned features of classifiers into well-defined political ideologies represented by news organizations.
We hope that the release of NewB will encourage further research on using natural language processing for political bias analysis.

\bibliography{emnlp2019}

\begin{thebibliography}{18}
\expandafter\ifx\csname natexlab\endcsname\relax\def\natexlab#1{#1}\fi

\bibitem[{Anuta et~al.(2017)Anuta, Churchin, and Luo}]{ElectionBias}
David Anuta, Josh Churchin, and Jiebo Luo. 2017.
\newblock \href {http://arxiv.org/abs/1701.06232} {Election bias: Comparing
  polls and twitter in the 2016 {U.S.} election}.
\newblock \emph{CoRR}, abs/1701.06232.

\bibitem[{Bernhardt et~al.(2008)Bernhardt, Krasa, and Polborn}]{MediaBias}
Dan Bernhardt, Stefan Krasa, and Mattias Polborn. 2008.
\newblock \href {https://doi.org/https://doi.org/10.1016/j.jpubeco.2008.01.006}
  {Political polarization and the electoral effects of media bias}.
\newblock \emph{Journal of Public Economics}, 92(5):1092 -- 1104.

\bibitem[{Cybenko(1989)}]{LogisticRegression}
George Cybenko. 1989.
\newblock \href {https://doi.org/10.1007/BF02551274} {Approximation by
  superpositions of a sigmoidal function. math cont sig syst (mcss) 2:303-314}.
\newblock \emph{Mathematics of Control, Signals, and Systems}, 2:303--314.

\bibitem[{Devlin et~al.(2018)Devlin, Chang, Lee, and Toutanova}]{BERT}
Jacob Devlin, Ming{-}Wei Chang, Kenton Lee, and Kristina Toutanova. 2018.
\newblock \href {http://arxiv.org/abs/1810.04805} {{BERT:} pre-training of deep
  bidirectional transformers for language understanding}.
\newblock \emph{CoRR}, abs/1810.04805.

\bibitem[{Gentzkow and Shapiro(2010)}]{MediaSlant}
Matthew Gentzkow and Jesse~M. Shapiro. 2010.
\newblock What drives media slant? evidence from us daily newspapers.
\newblock \emph{Econometrica}, 78:35--71.

\bibitem[{Gross et~al.(2013)Gross, Acree, Sim, and
  Smith}]{IdeologicalBookCorpus}
Justin Gross, Brice Acree, Yanchuan Sim, and Noah~A Smith. 2013.
\newblock Testing the etch-a-sketch hypothesis: A compu- tational analysis of
  mitt romney’s ideological makeover during the 2012 primary vs. general
  elections.
\newblock In \emph{APSA 2013 Annual Meeting Paper}.

\bibitem[{Iyyer et~al.(2014)Iyyer, Enns, Boyd-Graber, and
  Resnik}]{PolitIdeologyRNN}
Mohit Iyyer, Peter Enns, Jordan Boyd-Graber, and Philip Resnik. 2014.
\newblock Political ideology detection using recursive neural networks.
\newblock In \emph{Proceedings of the 52nd Annual Meeting of the Association
  for Computational Linguistics}, volume~1.

\bibitem[{Jacobson(2016)}]{DonaldDivides}
Gary~C. Jacobson. 2016.
\newblock Polarization, gridlock, and presidential campaign politics in 2016?
\newblock \emph{The ANNALS of the American Academy of Political and Social
  Science}, 667:226--246.

\bibitem[{Kim(2014)}]{CNN>RNN}
Yoon Kim. 2014.
\newblock \href {http://arxiv.org/abs/1408.5882} {Convolutional neural networks
  for sentence classification}.
\newblock \emph{CoRR}, abs/1408.5882.

\bibitem[{Liu et~al.(2016)Liu, Qiu, and Huang}]{LSTMRNN}
Pengfei Liu, Xipeng Qiu, and Xuanjing Huang. 2016.
\newblock \href {http://dl.acm.org/citation.cfm?id=3060832.3061023} {Recurrent
  neural network for text classification with multi-task learning}.
\newblock In \emph{Proceedings of the Twenty-Fifth International Joint
  Conference on Artificial Intelligence}, IJCAI'16, pages 2873--2879. AAAI
  Press.

\bibitem[{Morris(2007)}]{Media=Divide}
Jonathan~S. Morris. 2007.
\newblock \href {https://doi.org/10.1111/j.1540-6237.2007.00479.x} {Slanted
  objectivity? perceived media bias, cable news exposure, and political
  attitudes}.
\newblock \emph{Social Science Quarterly}, 88(3):707--728.

\bibitem[{Pang and Lee(2004)}]{SubjectivityDataset}
Bo~Pang and Lillian Lee. 2004.
\newblock A sentimental education: Sentiment analysis using subjectivity
  summarization based on minimum cuts.
\newblock In \emph{Proceedings of the 42nd ACL}, pages 271--278.

\bibitem[{Pennington et~al.(2014)Pennington, Socher, and Manning}]{GloVe}
Jeffrey Pennington, Richard Socher, and Christopher Manning. 2014.
\newblock Glove: Global vectors for word representation.
\newblock In \emph{Proceedings of the 2014 conference on empirical methods in
  natural language processing}.

\bibitem[{Pinholster(2016)}]{Pinholster1067}
Ginger Pinholster. 2016.
\newblock \href {https://doi.org/10.1126/science.352.6289.1067} {Journals and
  funders confront implicit bias in peer review}.
\newblock \emph{Science}, 352(6289):1067--1068.

\bibitem[{Rao and Spasojevic(2016)}]{MessageDemOrRep}
Adithya Rao and Nemanja Spasojevic. 2016.
\newblock Actionable and political text classification using word embeddings
  and lstm.
\newblock \emph{arXiv preprint}, arXiv:1607.02501.

\bibitem[{Rao et~al.(2010)Rao, Yarowsky, Shreevats, and
  Gupta}]{TwitUserOrientation}
Delip Rao, David Yarowsky, Abhishek Shreevats, and Manaswi Gupta. 2010.
\newblock Classifying latent user attributes in twitter.
\newblock In \emph{Proceedings of the 2nd international workshop on Search and
  mining user-generated contents}, pages 37--44.

\bibitem[{Thomas et~al.(2006)Thomas, Pang, and Lee}]{Convote}
Matt Thomas, Bo~Pang, and Lillian Lee. 2006.
\newblock Get out the vote: Determining support or opposition from
  congressional floor-debate transcripts.
\newblock In \emph{Proceedings of EMNLP}, pages 327--335.

\bibitem[{Tumasjan et~al.(2010)Tumasjan, Sprenger, Sandner, and
  Welpe}]{TwitElectionPred}
Andranik Tumasjan, Timm Sprenger, Philipp Sandner, and Isabell~M. Welpe. 2010.
\newblock Predicting elections with twitter: What 140 characters reveal about
  political sentiment.
\newblock \emph{ICWSM}.

\end{thebibliography}
\bibliographystyle{acl_natbib}

\end{document}